\documentclass[10pt,twocolumn,letterpaper]{article}

\usepackage[pagenumbers]{template_files/cvpr} 

\usepackage{epsfig}
\usepackage{graphicx}
\usepackage{soul}
\usepackage{bbm}
\setuldepth{foobar}

\usepackage{mathrsfs}
\usepackage{comment}
\usepackage{color}
\usepackage[table]{xcolor}
\usepackage{amsmath}
\usepackage{amssymb}
\usepackage{wrapfig}
\usepackage{multirow}
\usepackage[ruled]{algorithm2e} 
\usepackage{color}
\usepackage[utf8]{inputenc} 
\usepackage[T1]{fontenc}    
\usepackage{url}            
\usepackage{booktabs}       
\usepackage{amsfonts}       
\usepackage{nicefrac}       
\usepackage{microtype}      
\usepackage{xcolor}         
\usepackage{diagbox}

\definecolor{green}{rgb}{0, 0.5, 0}
\definecolor{orange}{rgb}{0.8, 0.6, 0.2}
\definecolor{red}{rgb}{1.0, 0.0, 0.0}
\definecolor{teal}{rgb}{0.0, 0.4, 0.4}
\definecolor{purple}{rgb}{0.65,0,0.65}
\definecolor{saffron}{rgb}{0.95,0.75,0.2}
\definecolor{turquoise}{rgb}{0.0,0.5,0.5}
\definecolor{black}{rgb}{0.0, 0.0, 0.0}
\definecolor{gray}{rgb}{0.5, 0.5, 0.5}

\AtEndPreamble{
    \usepackage[capitalize]{cleveref}
    \crefname{section}{section}{sections}
    \Crefname{section}{Section}{Sections}
    \Crefname{table}{Table}{Tables}
    \crefname{table}{table}{tables}
    \crefname{figure}{figure}{figures}
    \Crefname{figure}{Figure}{Figures}
    \crefname{equation}{}{}
    \Crefname{equation}{Eq.}{Eqs.}
}

\renewcommand{\paragraph}[1]{\vspace{.5em}\noindent\textbf{#1.}}

\newcommand{\rzz}[1]{{\color{black}#1}}

\newcommand{\AT}[1]{{\color[rgb]{.5,.5,1}{\bf [AT: #1]}}}

\newcommand{\TODO}[1]{\textbf{\color{red}[TODO: #1]}}
\newcommand{\lily}[1]{{\color{purple}#1}}
\newcommand{\Lily}[1]{{\textbf{\color{purple}Lily: #1}}}

\renewcommand{\TODO}[1]{}
\renewcommand{\Lily}[1]{}
\renewcommand{\lily}[1]{#1}

\newcommand{\ourModel}{\textsc{HiT}\xspace}

\newcommand{\pc}{\mathcal{X}}
\newcommand{\num}{M}
\newcommand{\levels}{L}
\newcommand{\level}{\ell}
\newcommand{\ppart}{P}
\newcommand{\pnum}{N}
\newcommand{\res}{R}
\newcommand{\voxgrid}{\mathbf{G}}
\newcommand{\channel}{D}
\newcommand{\feat}{\mathbf{Z}}
\newcommand{\code}{\mathbf{C}}
\newcommand{\parent}{\mathbf{p}}
\newcommand{\ponehot}{\tilde{\boldsymbol{p}}}
\newcommand{\psum}{\mathbf{s}} 
\newcommand{\sub}{\mathbf{s}}
\newcommand{\hnum}{H}

\newcommand{\attn}{\mathbf{A}}
\newcommand{\subs}{\mathcal{S}}

\newcommand{\queries}{\mathbf{Q}}
\newcommand{\keys}{\mathbf{K}}
\newcommand{\values}{\mathbf{V}}

\newcommand{\x}{\mathbf{x}}

\newcommand{\pocc}{\hat{\mathcal{O}}}
\newcommand{\occx}{\tilde{\mathcal{O}}}
\newcommand{\occ}{\mathcal{O}}

\newcommand{\loss}{\mathcal{L}}
\newcommand{\linear}{\mathcal{G}_{\phi}}
\newcommand{\convex}{\mathcal{C}}
\newcommand{\normal}{\mathbf{n}}
\newcommand{\offset}{\mathbf{o}}
\newcommand{\blend}{\delta}
\newcommand{\rot}{\mathbf{R}}
\newcommand{\angles}{\mathcal{E}}
\newcommand{\trans}{\mathbf{t}}
\newcommand{\scale}{\mathbf{s}}
\newcommand{\sdf}{\Phi}
\DeclareMathOperator*{\argmax}{arg\,max}

\newcommand{\tx}{\tilde{x}}

\usepackage{cancel}

\usepackage{enumitem}
\setlist[itemize]{noitemsep,leftmargin=*,topsep=0em}
\setlist[enumerate]{noitemsep,leftmargin=*,topsep=0em}

\definecolor{cvprblue}{rgb}{0.21,0.49,0.74}
\usepackage[pagebackref,breaklinks,colorlinks,citecolor=cvprblue]{hyperref}

\title{HiT: \underline{Hi}erarchical \underline{T}ransformers for Unsupervised 3D Shape Abstraction}
\author{
Aditya Vora\textsuperscript{1} \qquad
Lily Goli\textsuperscript{2} \qquad
Andrea Tagliasacchi\textsuperscript{1,2} \qquad
Hao Zhang\textsuperscript{1} \\
\\
\textsuperscript{1}Simon Fraser University \qquad 
\textsuperscript{2}University of Toronto \qquad 
}

\begin{document}


\twocolumn[{%
\renewcommand\twocolumn[1][]{#1}%
\maketitle
\begin{center}
    \centering
    \captionsetup{type=figure}
    \includegraphics[width=\textwidth]{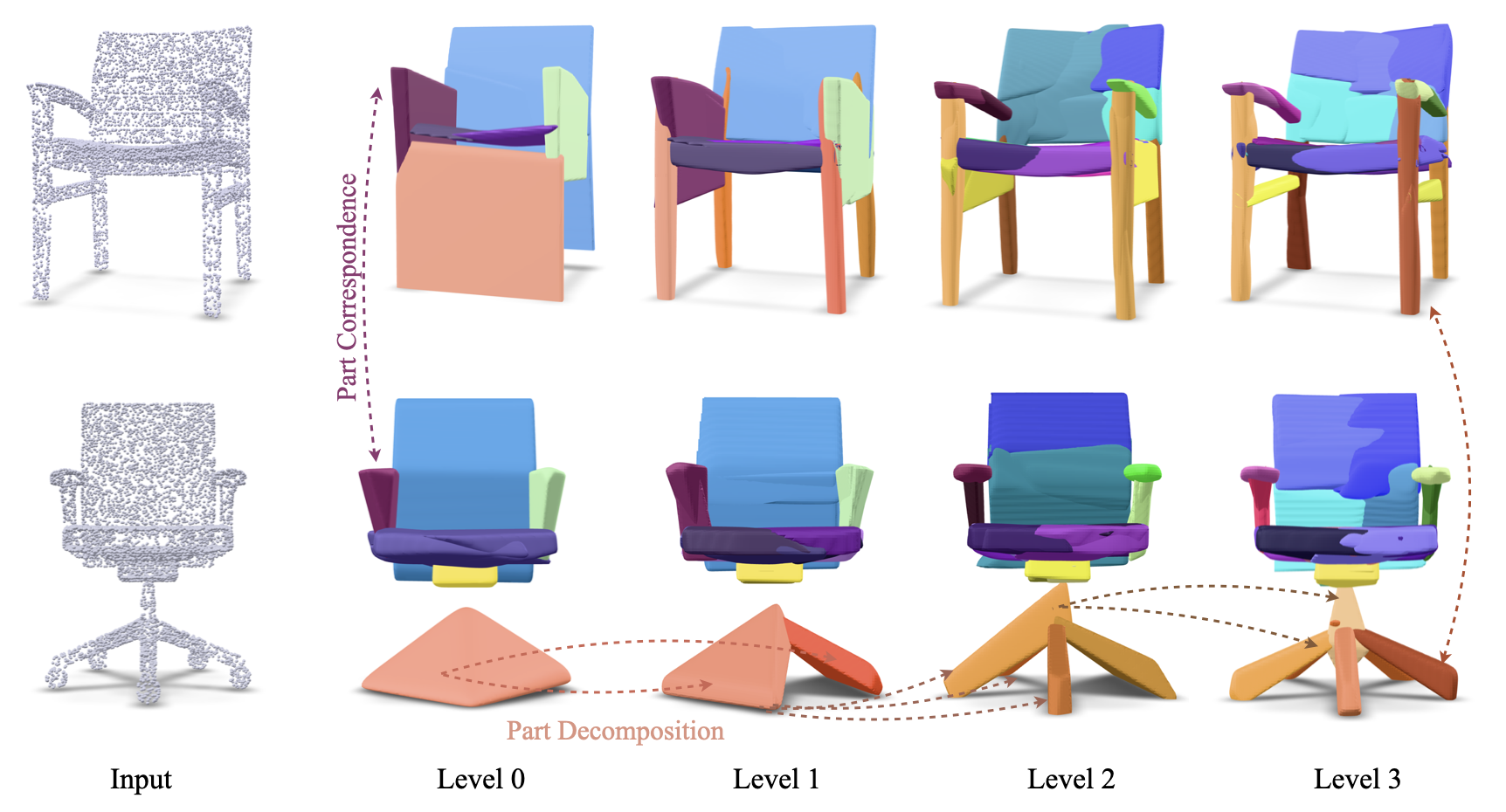}
        \caption{We present an attention-based architecture for hierarchical part abstraction of 3D objects.
        Our model flexibly adapts the number of parts, without supervision, allowing semantically similar regions (e.g. chair legs) to be decomposed, through the hierarchies, into different numbers of child parts depending on their geometry. While the resulting parts at finer levels of abstractions may no longer be semantic (e.g., split of the chair seats), part correspondences between shapes remain meaningful at different levels of the hierarchies.}
    \label{fig:teaser}
\end{center}%
}]

\maketitle

\begin{abstract}
We introduce \ourModel, a novel hierarchical neural field representation for 3D shapes that learns general hierarchies in a coarse-to-fine manner across different shape categories in an \ul{unsupervised} setting. Our key contribution is a hierarchical transformer (\ourModel), where each level learns parent–child relationships of the tree hierarchy using a compressed codebook. This codebook enables the network to automatically identify common substructures across potentially diverse shape categories. Unlike previous works that constrain the task to a fixed hierarchical structure (e.g., binary), we impose no such restriction, except for limiting the total number of nodes at each tree level.
This flexibility allows our method to infer the hierarchical structure directly from data, over multiple shape categories, and representing more general and complex hierarchies than prior approaches.
When trained at scale with a reconstruction loss, our model captures meaningful containment relationships between parent and child nodes.
We demonstrate its effectiveness through an unsupervised shape segmentation task over all 55 ShapeNet categories, where our method successfully segments shapes into multiple levels of granularity.
\end{abstract}
\\
\textit{Project Page:} \href{https://aditya-vora.github.io/HiT/}{\texttt{aditya-vora.github.io/HiT/}}

\section{Introduction}
\label{sec:introduction}
It is well established in cognitive science that humans perceive shapes as structured collections of parts, organized {\em hierarchically} \citep{palmer1977, hoffman1984}.
This hierarchical part cognition helps to reason about the function of each part~\citep{minsky1986}, and allows humans to effortlessly establish correspondences between similar parts in different shapes in a collection. 
Introducing similar hierarchical decompositions to 3D digital shapes not only aligns representations with human perception, but also enables key applications in computer graphics and robotics. 
In graphics, it facilitates part-aware editing~\citep{hertz2022spaghetti}, attribute transfer~\citep{yin2021_3DStyleNet}, compositional shape generation~\citep{egstar2020_struct}, and motion generation \cite{vora2025articulate, zhao2025advances}. In robotics, it can support affordance reasoning, manipulation planning~\citep{cpm2023}, and generalizable interaction with objects~\citep{song2025learning}, in a coarse-to-fine manner.

While several prior works~\citep{yu2019partnet, mo2019structurenet, li2017grass} achieve hierarchical part \rzz{learning} through direct supervision, their reliance on labeled 3D datasets limits the scalability and generalizability of these methods.
A practical alternative must be an unsupervised and generalizable representation trained across shapes, enabling part correspondences to emerge as in human perception.
Current unsupervised methods often frame \rzz{part representation learning} as shape reconstruction task using implicit part representations~\citep{deng2020cvxnet, chen2019bae, hertz2020pointgmm, niu2022rim}, with priors imposed via constrained part forms -- e.g., convex shapes~\citep{deng2020cvxnet} or low-capacity MLPs~\citep{chen2019bae}. 

While effective for single-level decompositions, multi-level extensions usually assume a fixed, often binary, tree structure~\citep{hertz2020pointgmm, niu2022rim}, which is \rzz{unnatural} and fails to capture the diversity of real-world object hierarchies. 
For instance, {number of chair legs can vary between instances}; see~\cref{fig:teaser}. Capturing this variability requires learning the hierarchy itself, and \textit{not} prescribing it.

Meanwhile, recent transformer-based models such as NeuMap~\citep{tang2023neumap} have effectively shown using attention to learn ``soft'' spatial correspondences in a flexible, data-driven way. 
These models suggest a promising path toward flexible, learnable structure, but they \textit{lack an explicit notion of hierarchy} or part-based abstraction.
This leaves open the question of how to combine the flexibility of attention-based models with the interpretability and structural grounding of hierarchical part \rzz{reasoning}. 
%
%

We propose \ourModel, a Hierarchical Transformer that performs \emph{multi-level} part decomposition by not only learning recurring parts across a shape collection at different levels of abstraction, but also \textit{modeling the relationships between parts} at successive levels of abstraction. 
At inference time, given an input shape in the form of a point cloud, \ourModel decomposes it into multiple parts at each level and dynamically assigns a tree structure tailored to that shape; see \cref{fig:teaser}.

Inspired by recent advances in transformer-based correspondence learning, \ourModel is built as a multi-layered transformer decoder, where each layer represents a set of parts using a learned codebook. 
Part–subpart relationships across levels are established via standard cross-attention, which softly assigns each subpart to a parent part at the level above. 
Each part is then mapped to a 3D convex primitive that provides a geometric description for that part, with subpart primitives encouraged to lie spatially within their assigned parent.

We show that \ourModel achieves state-of-the-art part decomposition performance on the ShapeNet/PartNet benchmarks, while producing \textit{geometrically interpretable} (as they are simply convex decompositions at each level) hierarchical abstractions of 3D shapes.
\section{Related work}
\label{sec:related}

Decomposing shapes into parts is a central problem in 3D understanding. We review methods that progress from unsupervised single-level segmentation to structured hierarchical representations, cover primitive-based implicit models for part reconstruction, and transformer-based approaches to hierarchical modeling in other domains. Finally, we review segmentation methods that utilize large pre-trained models.

\subsection{Structured neural shape representations}

\paragraph{Single level 3D shape structures}
A common approach to 3D shape understanding is to reconstruct them as compositions of primitives or semantic parts. Prior works learn such part-aware representations by approximating implicit surfaces with fixed or learned primitives. For example, \citep{paschalidou2020learning, tertikas2023generating, genova2019learning, hertz2020pointgmm, vora2023divinet, monnier2023differentiable} represent shapes as unions of superquadrics or Gaussians, but their fixed structures often fail to capture surfaces accurately.
BAE-Net~\citep{chen2019bae}, Neural-Parts~\citep{paschalidou2021neural}, DAE-Net~\citep{chen2024dae} and PartSDF ~\citep{talabot2025partsdf} improve flexibility using MLPs to represent parts, either through branching layers or by parameterizing deformations of simple shapes with invertible neural networks as learned homeomorphisms. While more expressive, these methods often yield coarse decompositions and can collapse without good initialization. 
\begin{figure*}[!t]
\centering
\includegraphics[height=5cm]
{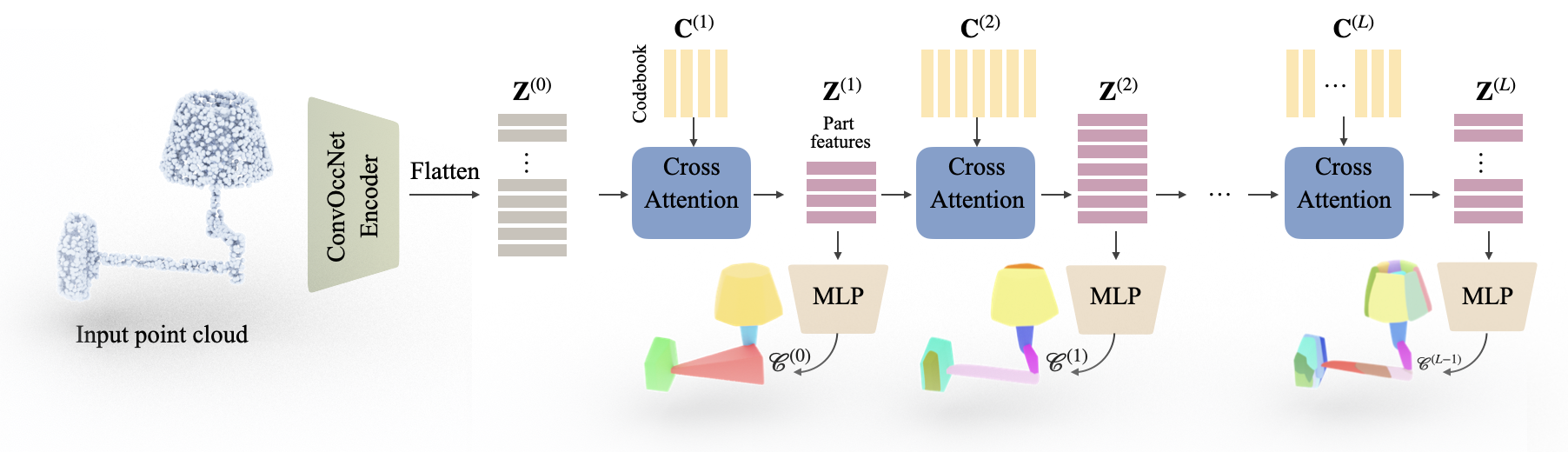}
\caption{ 
We propose a hierarchical transformer that learns part codebooks at each level, representing shapes from coarse to fine when trained across shapes.
Cross-attention ``connects'' levels, \textit{establishing learnable part–subpart relationships}.
The decoded parts are mapped to 3D convex primitives that provide geometric explanations.
An example decomposition of a lamp is shown across three levels, from a coarse base and shade to finer structural details.
}
\label{fig:pipeline}
\end{figure*}
CvxNet~\citep{deng2020cvxnet} and BSP-Net~\citep{chen2020bsp} balance expressiveness and precision by using convex primitives, which are differentiable and support fine-grained decomposition, but only at a single level.
Our approach extends this idea by learning a multi-level hierarchy of convex primitives with a transformer. Each level captures part composition and shared substructures, and a learned codebook enables reusable parts across shapes. This hierarchical design allows unsupervised part segmentation to emerge naturally from the representation.

\paragraph{Multi-level hierarchical 3D shape structure}
Part-whole hierarchies provide a deeper understanding of shape structure by modeling how fine parts group into larger components. GRASS~\citep{li2017grass} and PartNet~\citep{yu2019partnet} predict binary part trees using supervised recursive neural networks. StructureNet~\citep{mo2019structurenet} generalizes this to graph-based hierarchies, but still requires full supervision of part relationships. RIMNet~\citep{niu2022rim} takes an unsupervised approach to binary decomposition using implicit functions, while ~\citep{van2013co} explores co-hierarchical analysis of shape sets given pre-segmented parts. 
In contrast to all these methods, our model learns general n-ary hierarchies, without part annotations, and scales to diverse shape categories, discovering reusable substructures and coarse-to-fine part groupings directly from shape geometry.

\paragraph{Transformers for hierarchical modeling}
Transformers have recently been adapted to capture structural and hierarchical information across various domains. GLOM~\citep{hinton2023represent} introduced the idea of using a capsule-like transformer for hierarchically modeling part–whole relationships in compositional visual entities. In natural language processing, \citep{nawrot2021hierarchical} employed hierarchical transformers for long-range sequence modeling, while in computer vision, models such as Swin Transformer~\citep{liu2021swin} leverage hierarchical attention to capture multiscale spatial dependencies for object representation learning tasks.
Building on these lines of work, we propose a hierarchical transformer architecture specifically designed for learning structured representations of 3D shapes. Each level of our transformer captures parent–child relationships between parts through a shared codebook, enabling hierarchical reasoning and compositional generalization across shape categories.

\subsection{Shape segmentation with pre-trained models}
\lily{
Several recent methods infer 3D segmentation by importing information external to raw geometry. Multi‑view neural‑field approaches lift 2D semantic or panoptic masks from posed images into an implicit 3D volume, achieving scene‑level labels without part annotations~\citep{zhi2021place,vora2021nesf,tschernezki2022neural,kundu2022panoptic,siddiqui2023panoptic,hong20233d, fedele2025superdec}. Complementary work leverages vision-language training to segment shapes under zero‑shot or open‑vocabulary setups, from point‑cloud labeling~\citep{michele2021generative,chen2022zero,koo2022partglot, perla2025asia} to CLIP‑based mesh highlighting~\citep{ding2023pla,decatur20233d,abdelreheem2023satr}. All of these methods require either 2D mask supervision or large language–vision models, whereas our method learns part hierarchies directly from geometry without any auxiliary cues.}


\section{Method}
\label{sec: method}
\begin{figure*}[t]
\includegraphics[width=1.0\textwidth]{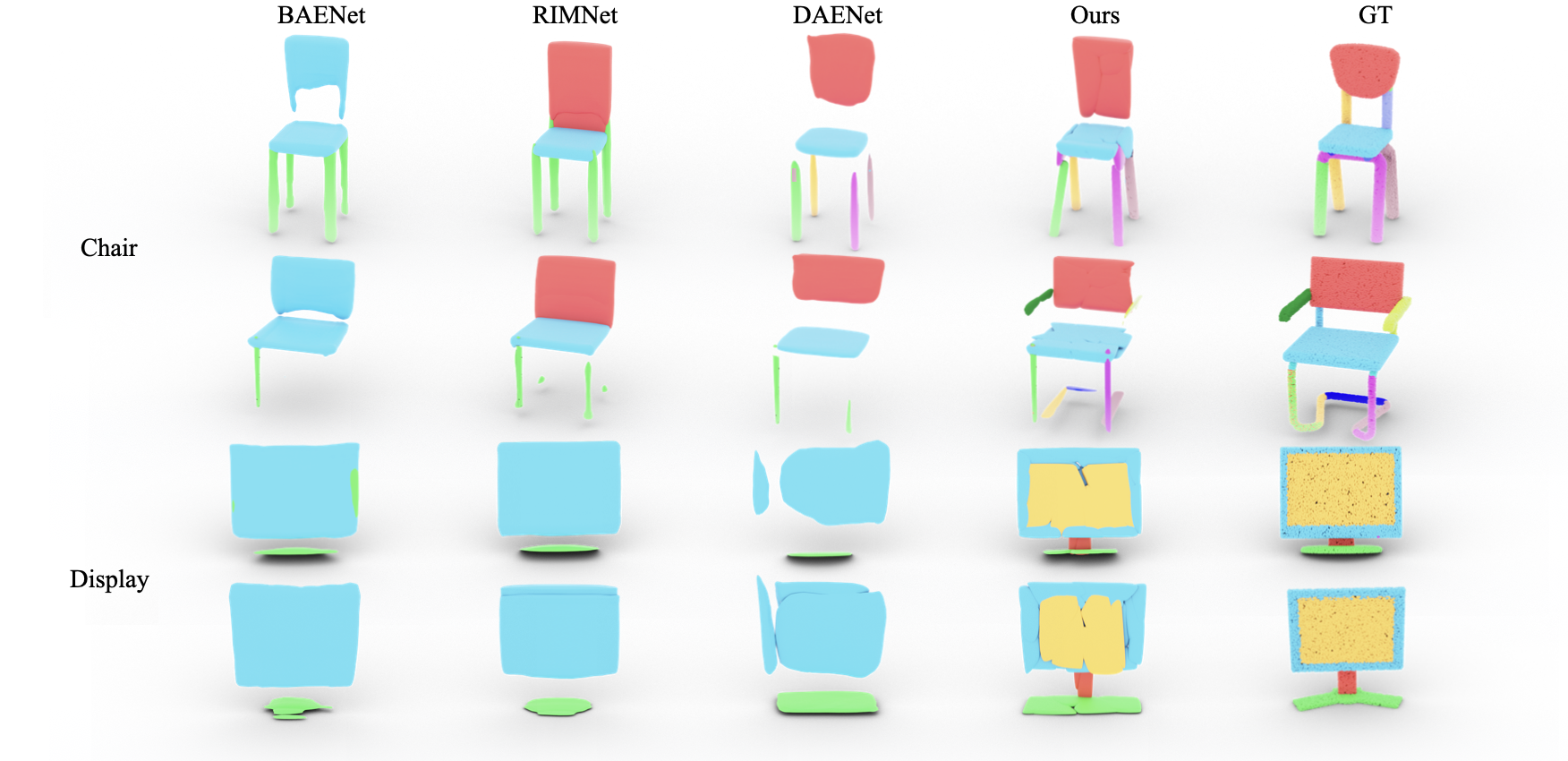}
\centering

\resizebox{\textwidth}{!}{
\begin{tabular}{c|cccccccccccc|c}
\toprule
\textbf{Method} & Bag & Bed & Bottle & Bowl & Chair & Clock & Washer & Disp. & Door & Earphone & Faucet & Hat & \textbf{Mean} \\
\midrule
BAENet~\cite{chen2019bae}     & 31.7 & 4.2 & 32.5 & 77.1 & 8.1 & 16.8 & 12.8 & 23.5 & 27.7 & 13.8 & 17.0 & 39.9 & 25.42 \\
RIMNet~\cite{niu2022rim}      & 40.3 & 9.1 & 56.2 & 83.3 & 18.0 & 21.2 & 20.7 & 37.8 & 29.5 & 21.4 & 25.6 & 53.9 & 34.75 \\
DAENet~\cite{chen2024dae}     & 43.7 & 12.6 & 58.1 & 82.1 & 31.2 & 24.2 & 28.9 & 41.9 & 32.7 & 34.3 & 36.4 & 55.4 & 40.28 \\
\cmidrule(lr){1-14}
\textbf{Ours} & \textbf{46.1} & \textbf{35.3} & \textbf{69.2} & \textbf{86.2} & \textbf{40.8} & \textbf{34.6} & \textbf{35.1} & \textbf{55.4} & \textbf{39.8} & \textbf{43.0} & \textbf{40.1} & \textbf{58.4} & \textbf{48.66} \\
\bottomrule
\end{tabular}
}

\vspace{-0.2cm}
\caption{ We outperform all baselines in the part segmentation task on ShapeNet, both qualitatively and quantitatively (IoU~$\uparrow$). Our dynamic tree structure adapts to geometry variations within a category (e.g., chairs), discovering a varying number of parts, while fixed-tree baselines fail to capture such differences.
} %
\label{fig:comparison_one}
\end{figure*}

%
Given an input point cloud $ \pc \in \mathbb{R}^{\num \times 3} $, our goal is to recover a \textit{hierarchical} decomposition of $\pc$ into disjoint \textit{parts}, across $ \levels $ levels.
We represent parts via 3D occupancy fields, and we require the decomposition to be semantically \textit{consistent} across shapes; see~\cref{fig:comparison_one,fig:ncomparison_two,fig:hierarchy}.
At each level $\level$, the shape is abstracted into a set of $\pnum_\level$ disjoint parts \scalebox{.77}{$\{\ppart^{(\level)}_p\}_{p=1}^{\pnum_\level}$}. Every parent part \scalebox{.77}{$\ppart^{(\level)}_p$} has a non-empty set of sub-parts \scalebox{.77}{$\subs_p^{(\level)}$} at level $\level+1$, such that,
\begin{equation}
     \bigcup_{\ppart^{(\level+1)}_s \in \subs_p^{(\level)}} 
     \left\{
     \ppart^{(\level+1)}_s 
     \right\}
     \;\; \subset \;\; \ppart^{(\level)}_p .
\end{equation}
Therefore, each parent is decomposed into sub-parts that are fully contained within it.
The number of parts $\pnum_\level$ is a hyper-parameter specified by the user, and it typically increases with $\level$, enabling progressively finer decompositions.
We impose \textit{no further constraints} on the structure of the hierarchy tree beyond the total number of parts at each level.
We propose a self-supervised encoder-decoder that discovers hierarchical part decompositions by reconstructing the occupancy field of shapes.
The resulting hierarchies emerge without supervision from part labels or predefined tree structures, and naturally adapt to shapes across categories.

We achieve this by introducing Hierarchical Transformer~(HiT) architecture, which features a $\levels$-layer decoder. 
Each layer represents one level of the hierarchy using a learnable codebook of $\pnum_\level$ parts.
To model hierarchical structure, cross-layer attention captures the relationships between parts and their subparts.
Each part is then grounded geometrically by mapping it to a 3D convex primitive, which is required to be fully contained within its parent.
The union of all convexes at a given level approximate the full shape.
In practice, HiT builds a \ul{differentiable tree structure}, where each node corresponds to a part embedding, part–subpart relationships are softly encoded in the cross-attention matrix, and each node is grounded by mapping to a localized convex region in 3D space; see~\cref{fig:teaser}.

\paragraph{Outline}
In~\cref{sec:architecture}, we describe how decoder layers use part codebooks and cross-attention to define part–subpart relationships.
In~\cref{sec: convexes}, we show how parts are geometrically grounded using 3D convex primitives with nested containment.
Finally,~\cref{sec:objectives} outlines the training objectives that combine reconstruction loss with regularizers to achieve self-supervised training.

\subsection{Hierarchical parts transformer}
\label{sec:architecture}
%
Our hierarchical part transformer consists of a standard point cloud encoder followed by $\levels$-layer part hierarchy decoder.
The decoder is built around two key decoder components:
\begin{enumerate}[label=(\roman*),leftmargin=*]
\item codebook-based information bottlenecks at each level, which learn semantically consistent part codes when trained on multi-category shape collections, and
\item a cross-attention mechanism that models part–subpart relationships across levels.
We now describe each component in detail.
\end{enumerate}

The architecture begins with a point cloud encoder adopted from ConvOccNet~\citep{peng2020convolutional}. 
The encoder maps each point to a $\channel$-dimensional latent feature, which are then pooled into a voxel grid $\voxgrid$ at a fixed resolution $\res$ using average pooling within each voxel. 
The resulting grid is flattened into a feature matrix ~$\feat^{(0)}$ with dimensions  ${\res^3 {\times} \channel}$, which serves as the input single-part representation at level zero of the $\levels$ layers deep HiT decoder hierarchy.

Each decoder level $\level$ contains a fixed-size learnable codebook $\code^{(\level)}$ of code parts that aim to capture recurring shape patterns at that level of abstraction.
To achieve this, the $\pnum_\level$ codes in the codebook act as queries into the incoming part features $\feat^{(\level -1)}$ from the previous level, enabling a \textit{soft assignment} of features to the parts represented by their codes:
\begin{equation}
\begin{gathered}
\feat^{(\level)} {=} \attn^{(\level)} \cdot \values^{(\level)}, \\
\attn^{(\level)} = \mathrm{SoftMax}\left( 
\tfrac{\queries^{(\level)} \cdot {\keys^{(\level)}}^\top}{\sqrt{\channel}} \right) ,
\\
\begin{cases}
\queries^{(\level)} {=} \mathbf{W}_\queries^{(\level)} \cdot \code^{(\level)}\\
\keys^{(\level)} {=} \mathbf{W}_\keys^{(\level)} \cdot \feat^{(\level-1)} \\
\values^{(\level)} {=} \mathbf{W}_\values^{(\level)} \cdot \feat^{(\level-1)}
\end{cases}
\end{gathered}
\end{equation}
This produces $\pnum_\level$ updated part features $\feat^{(\level)}$ for level $\level$.\footnote{Note that differently from a classical transformer where the number of tokens in the input matches the number of tokens in the output, in our architecture the number of tokens in the output layer matches the cardinality of the codebook within the layer.}
The benefit of this design is two-fold:
\begin{enumerate}[label=(\roman*)]
\item the codebooks act as information bottlenecks, encouraging the learned codes to capture recurring structures across shapes.
\item part–subpart relationships in the hierarchy are fully learnable and emerge via the attention matrix $\attn^{(\level)}$.
\end{enumerate}
The matrix $\attn^{(\level)}$ defines a soft adjacency between part features of the previous level and the part codes of the current level.
Hence, a single part $\parent$ from the previous level can be interpreted to be ``assigned'' as parent to each subpart $\sub$ at the current level as:
\begin{equation}
\parent = \argmax \attn^{(\level)}_{\sub, p}.
\label{eq:parent}
\end{equation}
To make this discrete parent selection differentiable, we use a straight-through estimator. Specifically, we define a pseudo one-hot vector $\ponehot {\in} \{0,1\}^{\pnum_{\level-1}}$, that behaves like a hard assignment in the forward pass but preserves gradients from the soft attention:
\begin{equation}
    \ponehot = \attn_{\sub, \cdot} + \cancel{\nabla}\left(\mathbbm{1}[\parent] - \attn_{\sub, \cdot}\right),
\label{eq:soft-parent}
\end{equation}
where $\cancel{\nabla}$ is the stop-gradient symbol, and $\mathbbm{1}[\parent]$ is a one-hot vector active at index $\parent$.

\subsection{Geometric part parametrization}
\label{sec: convexes}
With the tree hierarchy defined, we now describe how each recovered part is grounded to 3D geometry.
As the following applies identically across decoder levels, in what follows we drop the layer superscript. 
We take inspiration from CvxNet~\citep{deng2020cvxnet} in parameterizing each part as a 3D convex.

\paragraph{Representing parts as convexes}
At each level, we augment the decoder with $\linear$, a set of fully connected layers that map each subpart feature~$\feat_\sub$ to the parameters of a convex:
\begin{equation}
\convex_\sub = \linear(\feat_\sub).
\end{equation}
Each convex $\convex_\sub$ is defined by $\hnum$ half-spaces, parameterized by plane normals, offsets, and blending weights: $\{\normal^h_\sub, \offset^h_\sub, \blend_\sub\}_{h=1}^{\hnum}$.
In addition, each convex is assigned a rigid transformation specified by rotation parameters as Euler angles $\angles_\sub$, translation $\trans_\sub$, and scale $\scale_\sub$.
The occupancy field for subpart $\sub$ is then defined as:
\begin{equation}
\begin{gathered}
\occx_\sub(\x) = \mathrm{Sigmoid}(-\sigma \sdf_\sub(\x)), \\
\sdf_\sub(\x) = \log \sum_{h=1}^{\hnum} \exp\left(\normal_\sub^h \cdot \tx + \offset_\sub^h\right),
\\
\tx = \rot(\angles_\sub)^\top \left(\tfrac{\x - \trans_\sub}{\scale_\sub}\right)
\end{gathered}
\label{eq:convex_occ}
\end{equation}
where $\tx$ is the query point transformed into the local coordinate frame of the convex, and $\sigma$ is a hyperparameter controlling the sharpness of the SDF; We set $\sigma=75$.

\begin{figure*}[t]
\includegraphics[width=\textwidth]{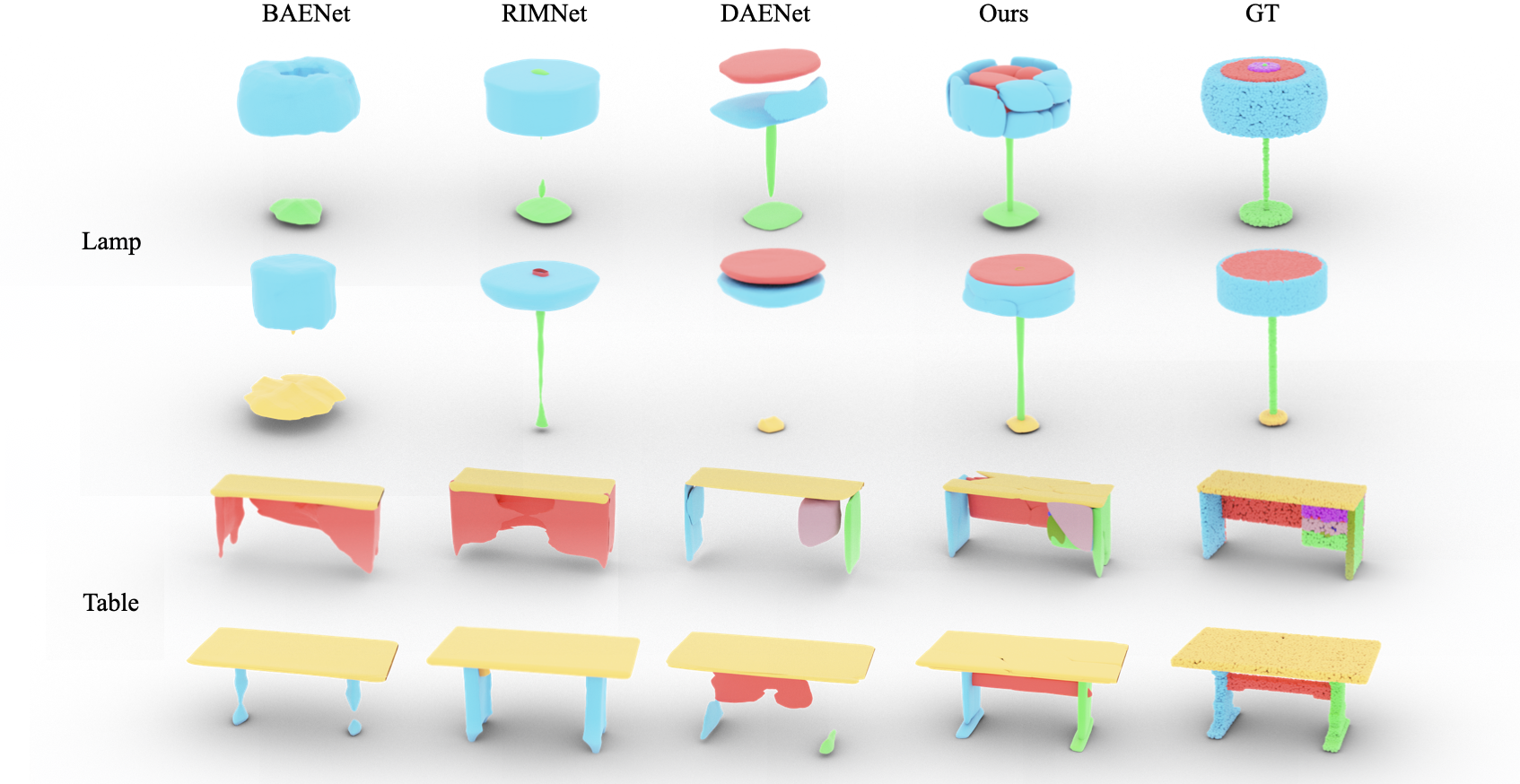}
\centering
\resizebox{\textwidth}{!}{
\begin{tabular}{c|cccccccccccc|c}
\toprule
\textbf{Method} & Key & Knife & Lamp & Laptop & Micro & Mug & Fridge & Scissors & Storage & Table & Trash & Vase & \textbf{Mean} \\
\midrule
BAENet~\cite{chen2019bae}     & 0.7 & 33.9 & 22.1 & 16.9 & 7.6  & 42.0 & 10.9 & 11.7 & 3.8  & 9.1  & 17.6 & 58.6 & 19.57 \\
RIMNet~\cite{niu2022rim}      & 0.8 & 54.4 & 35.6 & 18.6 & 9.1  & 42.4 & 13.3 & 16.2 & 11.3 & 15.9 & 29.5 & 68.4 & 26.29 \\
DAENet~\cite{chen2024dae}     & 1.0 & 28.1 & 38.4 & 17.7 & 12.9 & 56.2 & 25.7 & 17.1 & 21.7 & 30.1 & 33.1 & 65.9 & 28.98 \\
\cmidrule(lr){1-14}
\textbf{Ours} & \textbf{1.7} & \textbf{75.8} & \textbf{47.1} & \textbf{21.8} & \textbf{16.5} & \textbf{72.4} & \textbf{32.2} & \textbf{57.9} & \textbf{30.0} & \textbf{29.9} & \textbf{38.5} & \textbf{73.3} & \textbf{41.43} \\
\bottomrule
\end{tabular}
}

\vspace{-0.2cm}
\caption{
Qualitative and quantitative (IoU~$\uparrow$) results on the ShapeNet dataset show that our method achieves improved part segmentation by accurately reconstructing and consistently recovering recurring parts, whereas baselines often misclassify or miss them entirely.
} %
\label{fig:ncomparison_two}
\end{figure*}
\paragraph{Containment}
Although this formulation associates each subpart in the hierarchy with a geometric primitive, it does not alone enforce spatial consistency with the part–subpart relationships defined by the transformer.
To ensure spatial containment of sub-parts in parent parts, we modulate each sub-part’s occupancy with that of its parent:
\begin{equation}
\pocc_\sub(\x) =  \pocc_{\parent}(\x) \cdot \occx_\sub(\x), \quad \pocc_{\parent}(\x) = \sum_{p=0}^{\pnum_{\level -1}} \ponehot_p \cdot \pocc_p(\x) 
\label{eq:child-contain}
\end{equation}
where $\ponehot$ is the one-hot vector indicating the parent part for this subpart, defined by~\cref{eq:soft-parent}.
This constraint ensures that the subpart has valid nonzero occupancy only when its contained in its parent's spatial support.

\subsection{Training objectives}
\label{sec:objectives}
We train our network using a combination of losses: reconstruction of the shape’s occupancy field at each level, regularizations on the convex parameters, and structural constraints to maintain a balanced and spatially valid hierarchy:
\begin{equation}
\loss = \loss_{\text{recon}} + \lambda_1 \loss_{\text{contain}} + \lambda_2 \loss_{\text{cvxnet}} + \lambda_3 \loss_{\text{balance}}.
\end{equation}

\paragraph{Occupancy reconstruction}
The reconstruction loss is applied \textit{per-level}, encouraging the union of part occupancies to best approximate the ground truth occupancy at that level:
\begin{equation}
\loss_{\text{recon}} = \sum_{\level=0}^{\levels} \left( \occ(\x) - \max_p \{ \pocc_p^{(\level)}(\x) \}\right).
\end{equation}
While this, together with the containment constraint in~\cref{eq:child-contain}, encourages sub-parts to lie inside their parent in order to explain the shape occupancy, it does not strictly prevent sub-parts from ``bleeding'' outside their parent’s support.
To enforce containment, we define:
\begin{equation}
\loss_{\text{contain}} = \sum_{\level=0}^{\levels} \sum_{\sub=1}^{\pnum_{\level}} \left( 1 - \pocc_\parent(\x) \right) \cdot \occx_\sub(\x).
\end{equation}

\paragraph{Convex regularization}
For $\loss_{\text{cvxnet}}$, we adopt convex regularizers from CvxNet~\citep{deng2020cvxnet}.
Specifically, we use the \textit{decomposition} loss $\loss_{\text{decomp}}$ from~\cite[Eq. 4]{deng2020cvxnet} to discourage overlapping convexes that redundantly explain the same regions of the shape.
We also incorporate their \textit{guidance} loss and a slightly modified locality loss, $\loss_{\text{guide}}$ and $\loss_{\text{loc}}$, from~\cite[Eq. 6, 7]{deng2020cvxnet}, which discourage the formation of {``dead''} convexes~(i.e., those with near-zero volume and no contribution to shape reconstruction).
In particular, we modify $\loss_{\text{guide}}$ to make it a \textit{symmetric} (i.e.~two-sided) Chamfer distance, additionally minimizing the distance from each query point to its nearest convex center.
In combination with our containment constraint~\cref{eq:child-contain}, the locality and guidance losses prevent sub-parts from collapsing outside their parent.
Specifically, under~\cref{eq:child-contain}, any subpart lying entirely outside its parent will be assigned zero occupancy by construction, and therefore cannot satisfy these two objectives.
Thus, these losses help recover such sub-parts by pulling them back into a valid configuration.
\vspace{-0.15cm}

\paragraph{Balancing the tree}
We encourage more balanced tree structures, where each parent has a non-empty set of children.
We do this by minimizing the variance in the number of sub-parts assigned per parent~\citep{sun2021canonical}.
Letting attention columns represent soft assignments, we write this as:
\begin{equation}
\loss_{\text{balance}} = \sum_{\level=0}^{\levels} \sum_{p=1}^{\pnum_\level} \left( \psum_p - \frac{1}{\pnum_\level} \sum_q \psum_q \right)^2,
\\
\psum_p = \sum_{\sub=1}^{\pnum_{\level+1}} \attn_{\sub, p}.
\end{equation}

\begin{figure*}[t!]
\centering
\includegraphics[width=\textwidth]{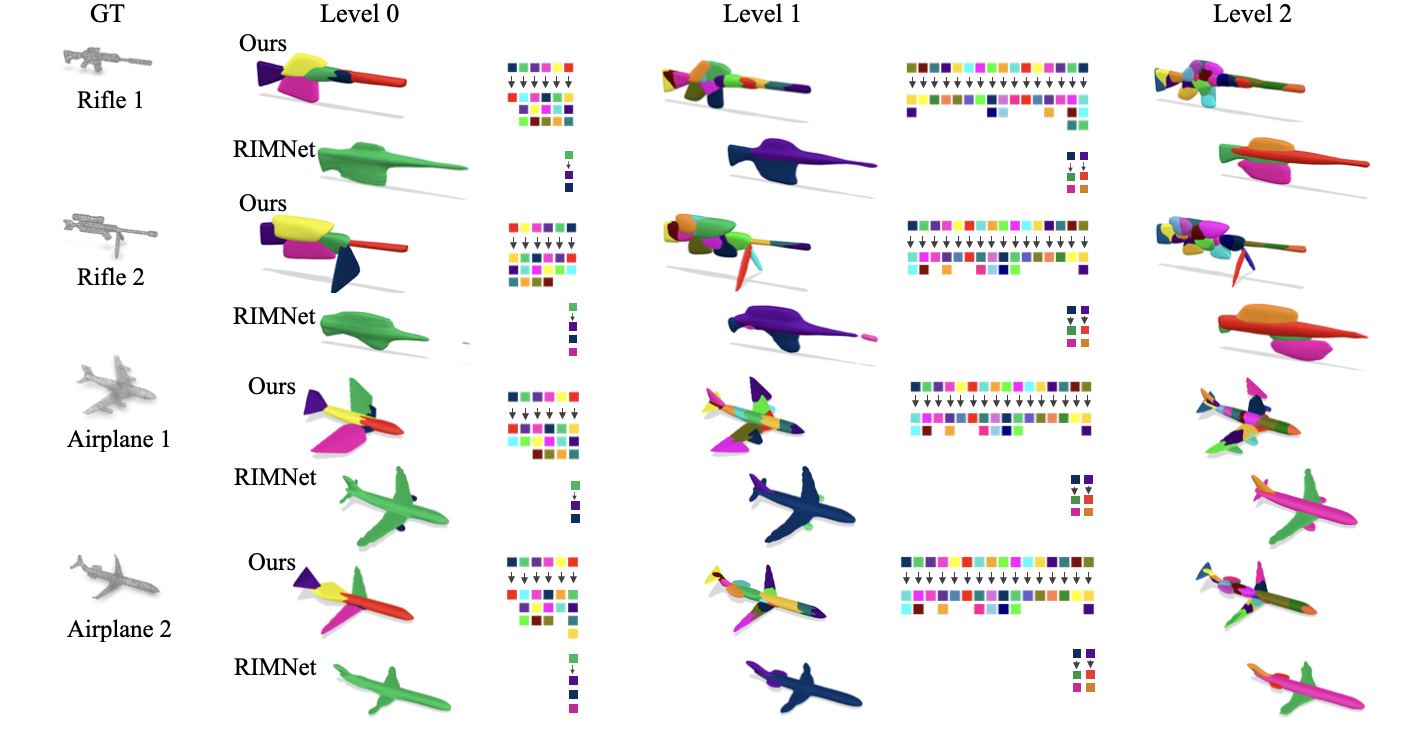}
\\[0.7em]
\setlength{\tabcolsep}{18pt}
\scriptsize
\resizebox{\textwidth}{!}{
\begin{tabular}{c|cccc|cccc}
\hline
\multirow{2}{*}{} & \multicolumn{4}{c|}{CD $(\downarrow)$}     & \multicolumn{4}{c}{IoU$(\uparrow)$}                 \\ \cline{2-9} 
                              & BAENet & DAENet & RIMNet & Ours            & BAENet & DAENet & RIMNet          & Ours            \\ \hline
Level 0                       & 0.0711 & 0.0643 & 0.0661 & \textbf{0.0630} & 0.6167 & 0.6154 & \textbf{0.6171} & 0.6023          \\
Level 1                       & -      & -      & 0.0658 & \textbf{0.0588} & -      & -      & \textbf{0.6192} & 0.6178          \\
Level 2                       & -      & -      & 0.0655 & \textbf{0.0576} & -      & -      & 0.6181          & \textbf{0.6242} \\
Level 3                       & -      & -      & 0.0663 & \textbf{0.0576} & -      & -      & 0.6214          & \textbf{0.6228} \\ \hline
\end{tabular}
}
\caption{
Our hierarchical part segmentation and reconstruction method produces a coherent multi-level shape abstraction: higher levels represent the main structural components, while finer levels capture detailed sub-parts, yielding more accurate reconstructions than prior approaches. The color maps show parent-child relationship between levels.
}
\label{fig:hierarchy}
\end{figure*}

\section{Experiments}
\label{sec: experiments}

We validate our method through the task of part segmentation~(\cref{sec:segmentaiton}), where we show our method can outperform all previous baselines in recovering consistent parts in all categories. We then show how our hierarchical part reconstruction achieves better reconstruction metrics in finer levels while higher levels remain semantically meaningful level of shape abstraction~(\cref{sec:recon}). Finally, we ablate our design choices and analyze our training dynamics and recovered codebooks~(\cref{sec:ablations}).


\subsection{Part segmentation -- \cref{fig:comparison_one,fig:ncomparison_two}}
\label{sec:segmentaiton}
We evaluate our hierarchical transformer on a common part segmentation benchmark and provide comparisons with state-of-the-art single- and multi-level part segmentation methods. 
We show a significant improvement in segmentation results, while recovering more coherent parts across each class of objects.

\paragraph{Dataset} We train our method on $55$ categories from the ShapeNet-v2 dataset, following the train-test split used by~\citet{zhang20233dshape2vecset}. 
Our approach supports a unified training scheme, allowing all categories to be trained jointly. We further analyze this design in \cref{sec:ablations}, where we compare it to category-specific training. For segmentation accuracy evaluation, since the ShapeNet \cite{chang2015shapenet} dataset does not provide segmentation labels, we leverage the \textit{fine-grained} part annotations from the PartNet \cite{mo2019partnet} dataset.
This differs from previous methods, which evaluate segmentation accuracy on the ShapeNetPart dataset \cite{yi2016scalable}, containing only \textit{coarse} segmentation ground-truth labels. 
The Door, Scissors, and Refrigerator categories appear only in PartNet and not in ShapeNet; hence for these, we use all instances for evaluation only.
Each ground-truth point cloud for evaluation in PartNet dataset contains $10,000$ points. 

\paragraph{Metrics} 
We evaluate segmentation performance using average Intersection over Union (IoU) on the segmented point cloud, following prior works~\citep{chen2019bae, chen2024dae, niu2022rim}. 
Each ground-truth point in the input point cloud is treated as a query to the predicted convexes at the final level of the hierarchy, and is assigned the label of the convex with the highest occupancy value.
Following BSPNet~\citep{chen2020bsp}, we assign a ground-truth-consistent label to each code in the codebook by identifying the label with the highest number of points falling into the code's corresponding convex. 
This label-code association is computed once per category on a single instance and then used for all other instances within that category. 

\paragraph{Baselines} 
We compare against BAENet~\citep{chen2019bae}, RIMNet~\citep{niu2022rim}, and DAENet~\citep{chen2024dae} for part segmentation. 
BAENet and DAENet perform single-level segmentation, while RIMNet uses a fixed binary hierarchy. 
For evaluation, we use the predicted parts from the final level of each method. All baselines require voxelized occupancy as input; therefore, we use a CUDA-accelerated voxelizer (binvox) to voxelize the watertight meshes provided in~\citep{zhang20233dshape2vecset}. For fair comparison, we expand the output branching factor of all baselines to $32$ to match the number of leaf nodes to our network and train them accordingly. 
In particular, we train RIMNet for $5$ levels to match the total number of leaf nodes with our model.

\paragraph{Analysis}
Our method significantly outperforms all baselines in part segmentation, both quantitatively and qualitatively as shown in Fig. \ref{fig:comparison_one} and \ref{fig:ncomparison_two}. 
It produces consistent labeling of corresponding parts across instances within a category, whereas baselines often miss parts entirely or misclassify semantically similar regions. 

\begin{figure}
\centering
\includegraphics[width=\linewidth]{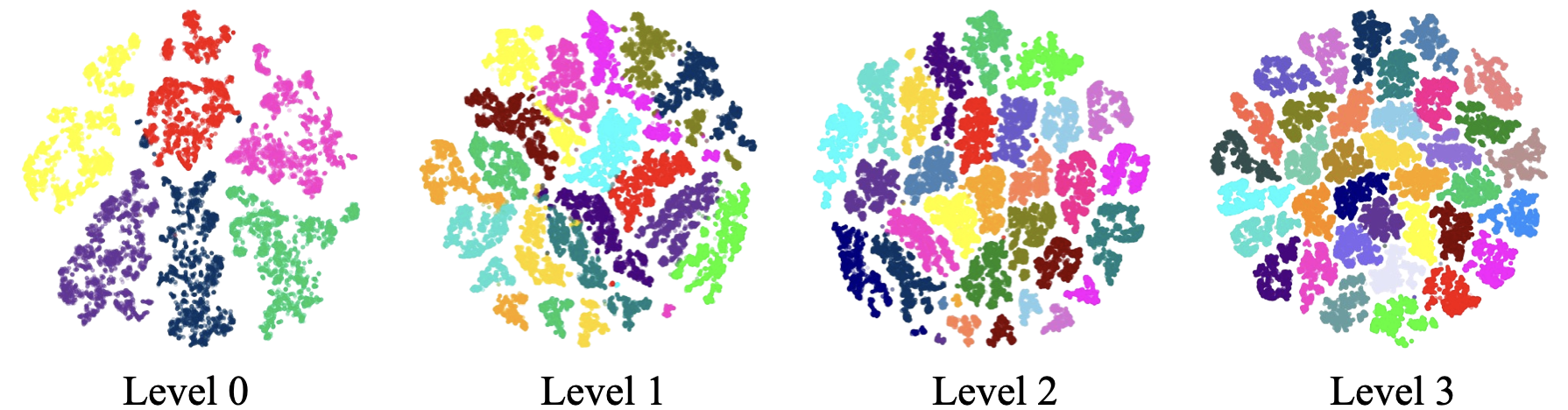}
\caption{t-SNE visualization of subpart features 
$\feat^{(\level)}$
  across the ShapeNet test set shows that embeddings for each part (color-coded) form coherent clusters in the embedding space.}
\label{fig:tsne}

\vspace{-0.3cm}
\end{figure}

\subsection{Hierarchical reconstruction -- \cref{fig:hierarchy}}
\label{sec:recon}
We demonstrate that our model captures coherent high-level abstractions at early levels and progressively refines fine-grained details at deeper levels. Following the protocol of \cite{zhang20233dshape2vecset}, we train and test on ShapeNet-v2 using the official splits. For evaluation, we compare reconstructed meshes—sampled to $100{,}000$ points each—against ground-truth point clouds, using symmetric Chamfer Distance and voxelized IoU at $128^{3}$ resolution. Our hierarchy employs $[6,16,24,36]$ parts. We compare against BAENet~\citep{chen2019bae}, RIMNet~\citep{niu2022rim}, and DAENet~\citep{chen2024dae}.
While BAENet and DAENet are single-level methods, RIMNet supports hierarchical reconstruction but only via a fixed binary tree, which yields coarse abstractions and loss of detail in later levels (\cref{fig:hierarchy}).
By contrast, our model recovers flexible multi-branch hierarchies that balance abstraction and detail across levels.

\begin{figure}
\includegraphics[width=\columnwidth]{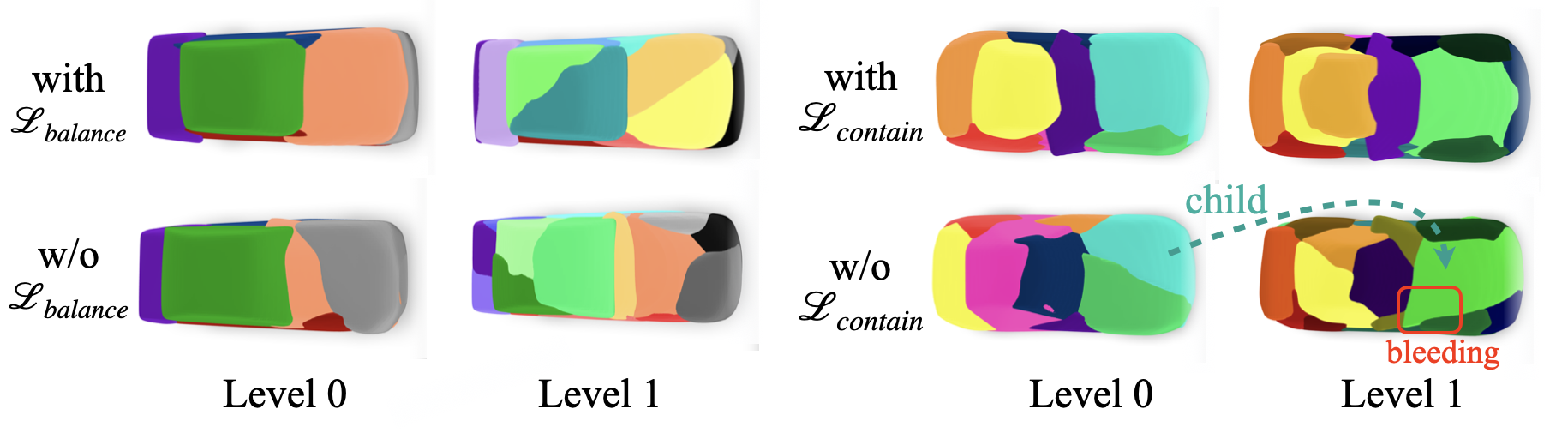}
\centering
\resizebox{\columnwidth}{!}{
\begin{tabular}{c|cccccc}
  \toprule
    \diagbox{Metric}{Loss}  & All & $-\mathcal{L}_{decomp}$ & $-\mathcal{L}_{loc}$ & $-\mathcal{L}_{guide}$ & $-\mathcal{L}_{contain}$ & $-\mathcal{L}_{balance}$ \\
\midrule
Vol-IoU ($\uparrow)$ & \textbf{0.723} & $0.669$  & $0.671$   & $0.662$  & $0.719$ & $0.721$  \\
CD ($\downarrow)$ & \textbf{0.044} & $0.063$ & $0.069$  & $0.065$ & \textbf{0.044} & $0.045$  \\
\bottomrule
\end{tabular}
}
\caption{
We show the effect of each of our loss components on part segmentation. (Left) Balance loss helps achieving a more even decomposition. (Right) Addition of containment loss prevents child parts bleeding out of their parent convex. }%
\label{fig:loss_ablations}
\vspace{-0.5cm}
\end{figure}

\begin{figure}[htbp]
\centering
\includegraphics[width=\columnwidth]{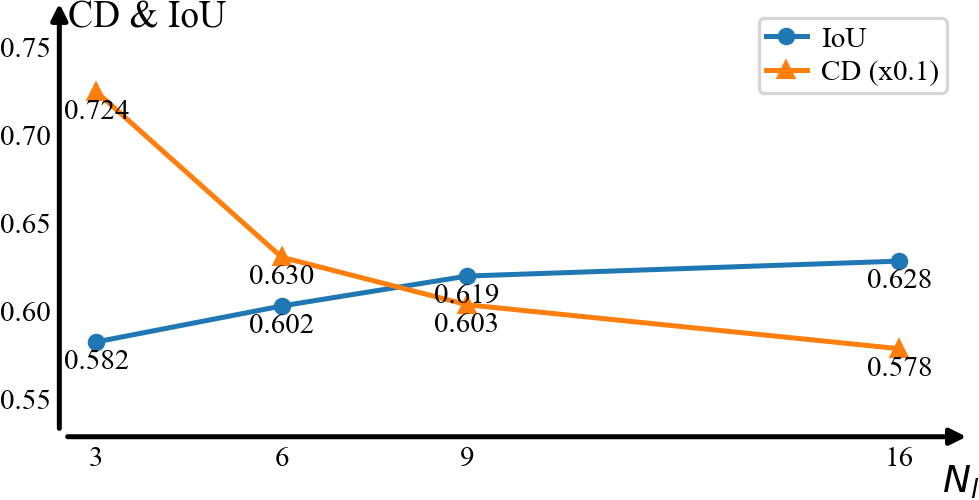}
\vspace{-0.5cm}
\caption{
CD and IoU comparison with varying number of parts. As number of parts increases the reconstruction quality improves. $N_{l}$ is the number of parts at level $0$. CD values are scaled by $0.1$.
}
\label{fig:n_parts_analysis}
\end{figure}

\subsection{Ablation study -- \cref{tab:cd_iou,fig:loss_ablations,fig:tsne,fig:n_parts_analysis}}
\label{sec:ablations}




We ablate our design choices by analyzing the effect of each loss, showing all loss components contribute to the quality of the decomposition. While balance and containment losses have little effect on quantitative metrics, removing them yields imbalanced and non–self-contained hierarchies, as seen in the qualitative results.
While our method offers an effective, convenient, and generalized training on all object categories at once, we further analyze how a per-category training similar to previous works \cite{chen2019bae, niu2022rim, chen2024dae} affects our results. This shows that a per-category training can result in an improved specialized model for each category, in exchange for more time and compute. We provide a t-SNE analysis of the learned part embeddings, demonstrating that embeddings of similarly labeled parts cluster closely together. Finally, in \cref{fig:n_parts_analysis} we show that reconstruction quality improves as number of parts increases.  

\begin{table}[t]
\centering
\footnotesize
\setlength{\tabcolsep}{2pt}
{
\begin{tabular}{l|l|cccccc|c}
\hline
\multirow{2}{*}{\textbf{Metric}} & \multirow{2}{*}{\textbf{Method}} & \multicolumn{6}{c|}{\textbf{Categories}} & \multirow{2}{*}{\textbf{Mean}} \\ \cline{3-8}
                                 &                                   & Bed & Chair & Display & Faucet & Earphone & Lamp  \\ \hline
\multirow{4}{*}{IoU $(\uparrow)$}
  & BAENet     & $5.2$ & $14.9$ & $36.2$ & $16.5$ & $16.7$  & $28.9$ & $19.73$ \\
  & RIMNet     & $7.3$ & $19.2$ & $44.8$ & $25.5$ & $16.8$  & $39.7$ & $25.55$ \\
  & DAENet     & $14.1$ & $29.9$ & $49.7$ & $37.9$ & $30.4$  & $38.3$ & $33.38$ \\
  \cline{2-9}
  & \textbf{Ours} & $\textbf{34.6}$ & $\textbf{38.8}$ & $\textbf{56.2}$ & $\textbf{41.3}$ & $\textbf{42.7}$ & $\textbf{45.3}$ & $\textbf{43.15}$ \\ \hline
\end{tabular}
}
\caption{
Our model can also be trained per-category, as opposed to on all shapes. We show results across six ShapeNet categories \citep{chang2015shapenet}. 
}
\label{tab:cd_iou}
\vspace{-0.3cm}
\end{table}

\section{Conclusion}
\label{sec: conclusion}

We introduce \ourModel, a self-supervised attention-based hierarchical neural field representation that learns general shape abstractions in a coarse-to-fine manner across diverse categories.
Our core contribution is a novel cross-attention mechanism which enables dynamic discovery of parent-child relationships across hierarchy levels. To the best of our knowledge, our method is the first to enable general cross-category hierarchical 3D shape abstraction.

However, our current method is constrained by the requirement that the number of parts at each hierarchy level must be fixed. Such a constraint can sometimes lead to unnatural decompositions, particularly at finer levels of the hierarchy, limiting its effectiveness in downstream applications. Recent trends that leverage foundation models as semantic priors offer a promising way to address this limitation, as they enable semantically meaningful grouping of primitives based on high-level prior knowledge. 
A promising future direction is to make the number of convexes adaptive, for example by selecting them based on sparsity or reconstruction quality, so that the hierarchy remains compact and semantically meaningful. 

\noindent \textbf{Acknowledgements.} We thank all the anonymous reviewers for their insightful comments. This research was supported by Adobe gift funds. \par

{
    \small
    \bibliographystyle{template_files/ieeenat_fullname}
    \bibliography{main}
}

\appendix
\renewcommand{\thesection}{\Alph{section}}
\setcounter{section}{0}

\section{Supplementary}
\vspace{1.2cm}

\begin{figure*}[t]
\centering
    \includegraphics[scale=0.75]
    {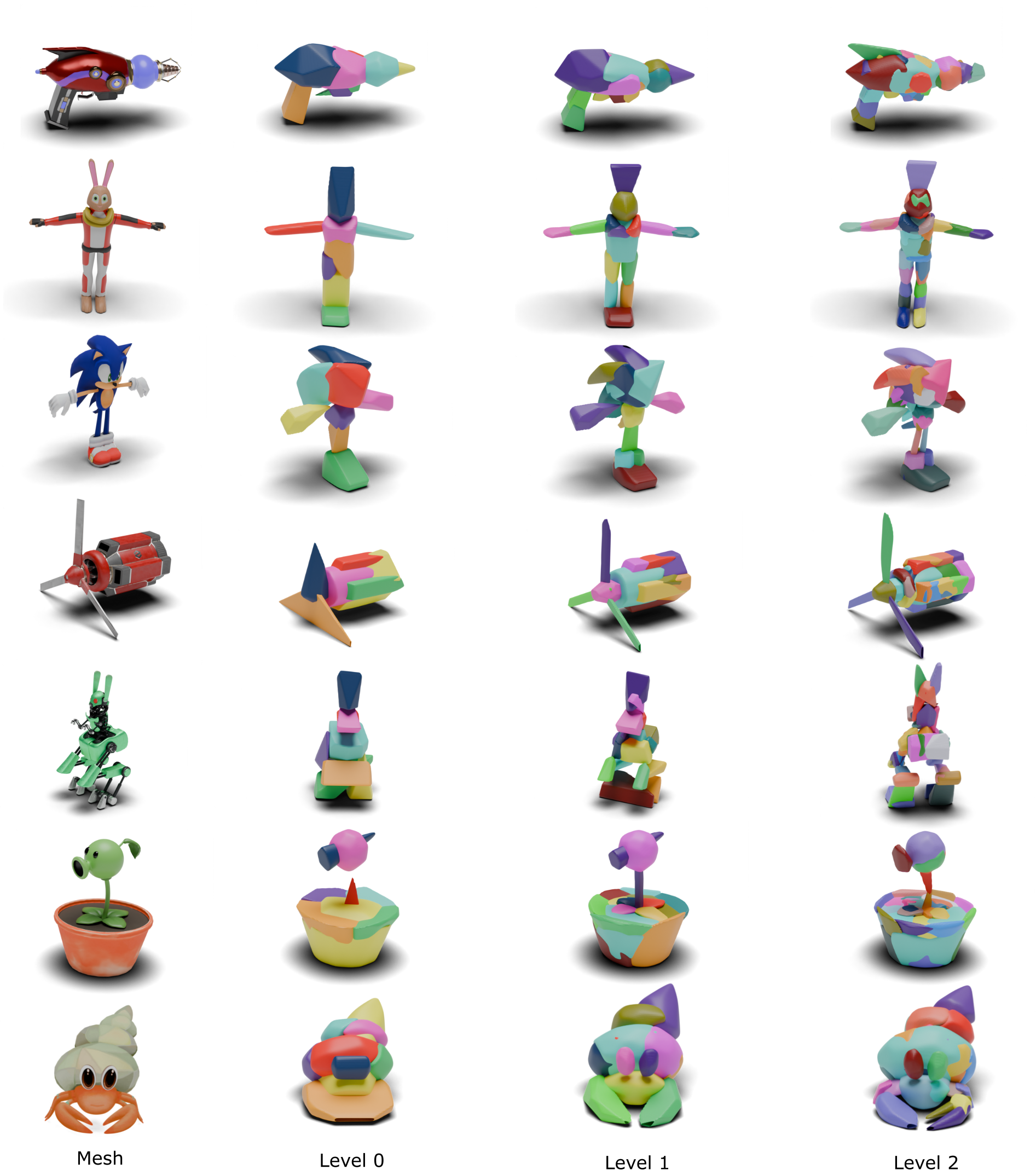}
\caption{Hierarchical decomposition results on Objaverse dataset ~\citep{deitke2023objaverse}. First column indicates the ground truth mesh from which points are sampled for our network input. Next 3 columns indicate hierarchical decomposition at multiple granularities predicted by our model.}
\label{fig:supp_obj_v3}
\end{figure*}

\begin{figure*}[t]
\centering
    \includegraphics[scale=0.7]
    {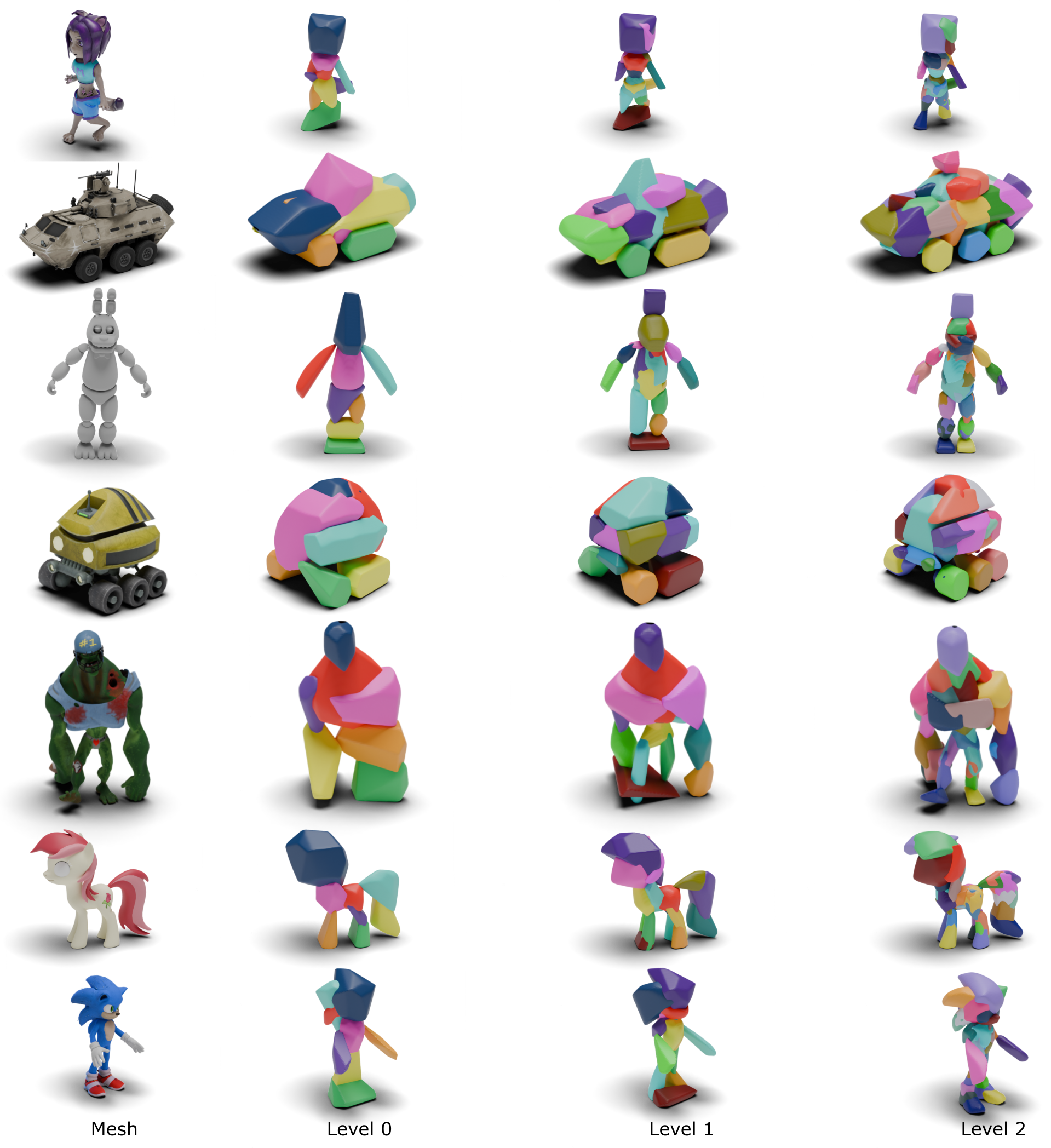}
\caption{Hierarchical decomposition results on Objaverse dataset ~\citep{deitke2023objaverse}. First column indicates the ground truth mesh from which points are sampled for our network input. Next 3 columns indicate hierarchical decomposition at multiple granularities predicted by our model.}
\label{fig:supp_obj_v4}
\end{figure*}

\begin{figure*}[t]
\centering
    \includegraphics[scale=0.7]{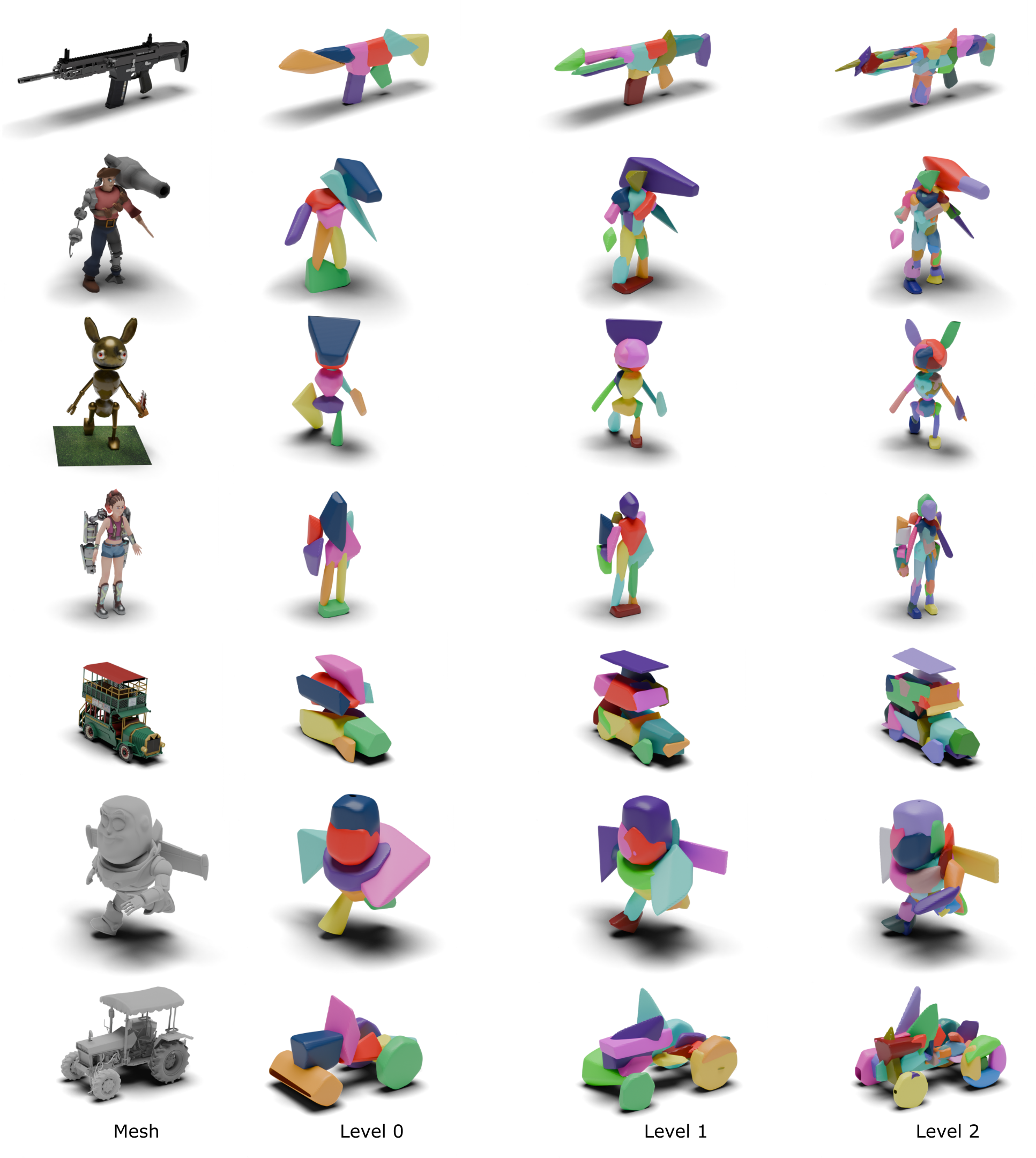}
\caption{Hierarchical decomposition results on Objaverse dataset ~\citep{deitke2023objaverse}. First column indicates the ground truth mesh from which points are sampled for our network input. Next 3 columns indicate hierarchical decomposition at multiple granularities predicted by our model.}
\label{fig:supp_obj_v1}
\end{figure*}

\begin{figure*}[t]
\centering
    \includegraphics[scale=0.75]{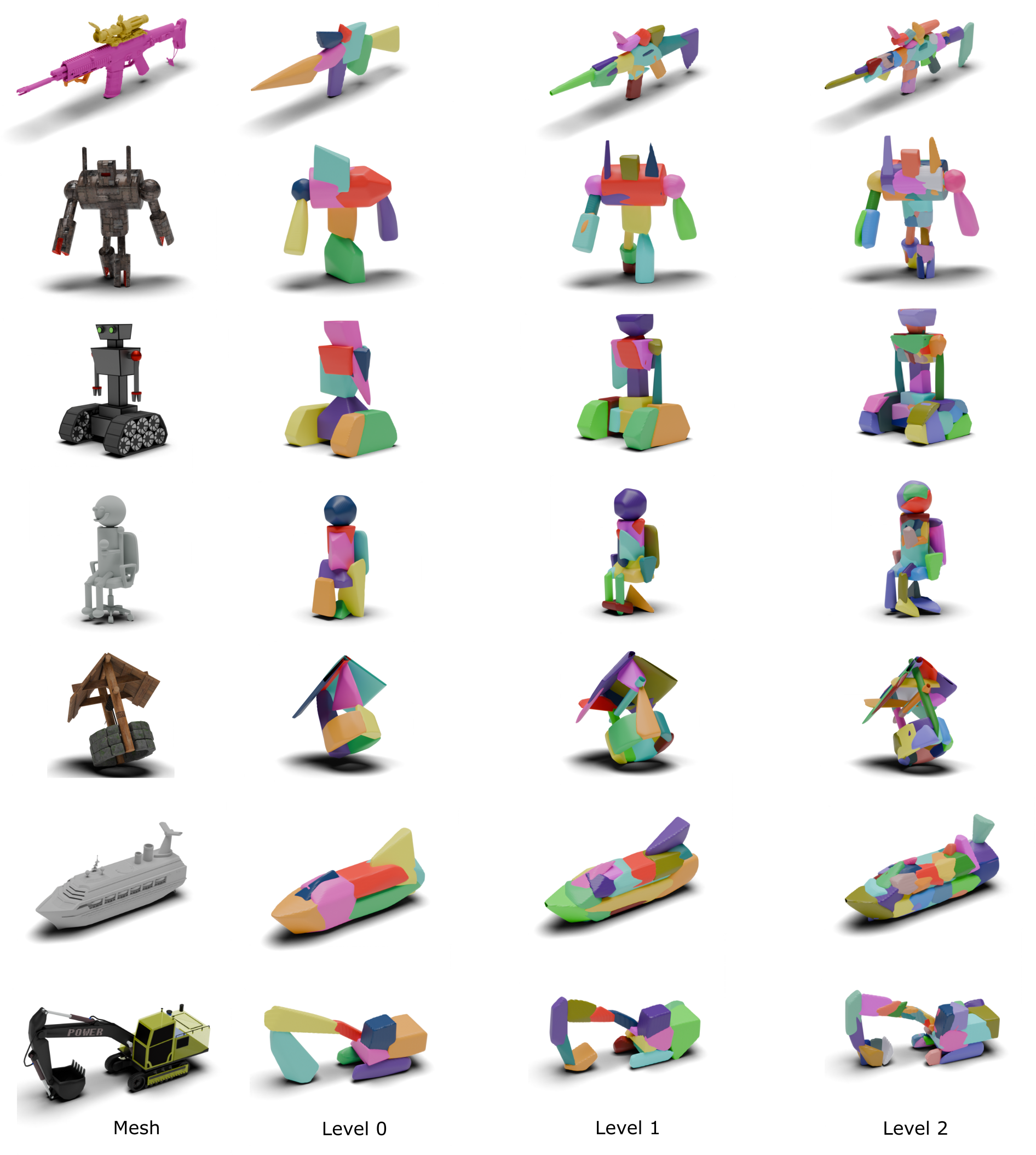}
\caption{Hierarchical decomposition results on Objaverse dataset ~\citep{deitke2023objaverse}. First column indicates the ground truth mesh from which points are sampled for our network input. Next 3 columns indicate hierarchical decomposition at multiple granularities predicted by our model.}
\label{fig:supp_obj_v2}
\end{figure*}

\begin{figure*}[t]
\centering
    \includegraphics[scale=0.7]{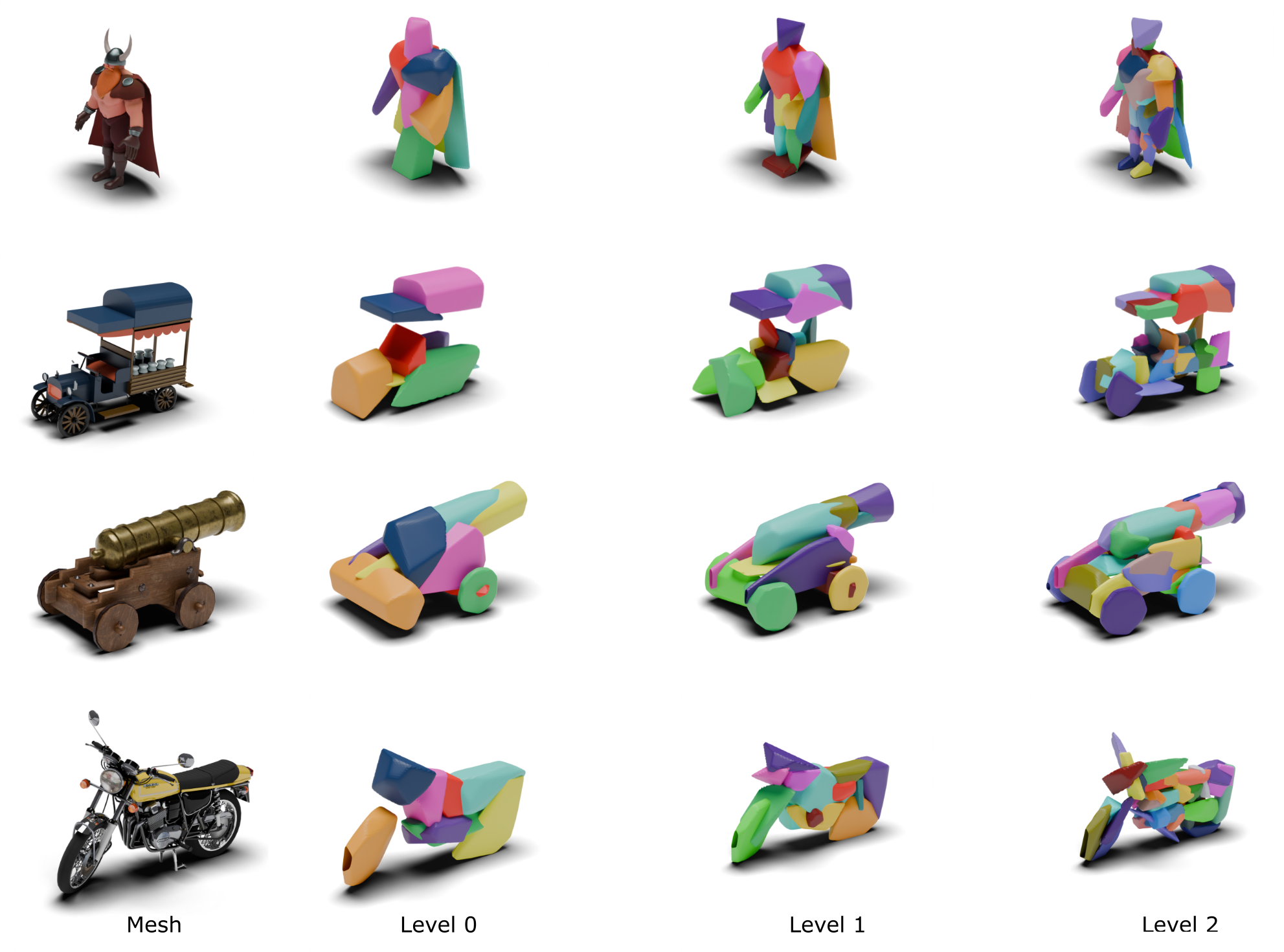}
\caption{Hierarchical decomposition results on Objaverse dataset ~\citep{deitke2023objaverse}. First column indicates the ground truth mesh from which points are sampled for our network input. Next 3 columns indicate hierarchical decomposition at multiple granularities predicted by our model.}
\label{fig:supp_obj_v5}
\end{figure*}

In the supplementary material, we present the hierarchical decomposition results of our method on the Objaverse dataset \cite{deitke2023objaverse}. As shown in the examples below, our model dynamically allocates a different number of convexes depending on the input shape. Figures \ref{fig:supp_obj_v3}, \ref{fig:supp_obj_v4}, \ref{fig:supp_obj_v1}, \ref{fig:supp_obj_v2}, and \ref{fig:supp_obj_v5} provide qualitative results of these decompositions. As can be seen, our model generates diverse decompositions tailored to the structural characteristics of each shape.

\end{document}